\documentclass[screen,nonacm]{acmart}

\setcopyright{none}
\renewcommand\footnotetextcopyrightpermission[1]{}
\usepackage{subfigure}
\usepackage{multirow}
\usepackage{nicematrix}

\AtBeginDocument{%
  \providecommand\BibTeX{{%
    \normalfont B\kern-0.5em{\scshape i\kern-0.25em b}\kern-0.8em\TeX}}}

\begin{document}

\title{Recent advances in interpretable machine learning using structure-based protein representations}

\author{Luiz Felipe Vecchietti}
\email{lfelipesv@ibs.re.kr}
\orcid{0000-0003-2862-6200}
\affiliation{%
  \institution{Institute for Basic Science}
  \city{Daejeon}
  \country{South Korea}
}

\author{Minji Lee}
\affiliation{%
  \institution{Korea Advanced Institute of Science and Technology}
  \city{Daejeon}
  \country{South Korea}}
\email{haewon_lee@kaist.ac.kr}
\orcid{0000-0001-5293-8904}

\author{Begench Hangeldiyev}
\affiliation{%
  \institution{Korea Advanced Institute of Science and Technology}
  \city{Daejeon}
  \country{South Korea}}
\email{begahan@kaist.ac.kr}
\orcid{0009-0008-2614-8843}

\author{Hyunkyu Jung}
\affiliation{%
  \institution{Korea Advanced Institute of Science and Technology}
  \city{Daejeon}
  \country{South Korea}}
\email{dino8egg@kaist.ac.kr}
\orcid{0009-0005-3725-4092}

\author{Hahnbeom Park}
\affiliation{%
  \institution{Korea Institute of Science and Technology}
  \city{Seoul}
  \country{South Korea}}
\email{hahnbeom@kist.re.kr}
\orcid{0000-0002-7129-1912}

\author{Tae-Kyun Kim}
\affiliation{%
  \institution{Korea Advanced Institute of Science and Technology}
  \city{Daejeon}
  \country{South Korea}}
\email{kimtaekyun@kaist.ac.kr}
\orcid{0000-0002-7587-6053}

\author{Meeyoung Cha}
\affiliation{%
  \institution{Max Planck Institute for Security and Privacy}
  \city{Bochum}
  \country{Germany}}
\email{mia.cha@mpi-sp.org}
\orcid{0000-0003-4085-9648}

\author{Ho Min Kim}
\affiliation{%
  \institution{Korea Advanced Institute of Science and Technology}
  \city{Daejeon}
  \country{South Korea}}
\email{hm_kim@kaist.ac.kr}
\orcid{0000-0003-0029-3643}

\renewcommand{\shortauthors}{}

\begin{abstract}
Recent advancements in machine learning (ML) are transforming the field of structural biology. For example, AlphaFold, a groundbreaking neural network for protein structure prediction, has been widely adopted by researchers. The availability of easy-to-use interfaces and interpretable outcomes from the neural network architecture, such as the confidence scores used to color the predicted structures, have made AlphaFold accessible even to non-ML experts. In this paper, we present various methods for representing protein 3D structures from low- to high-resolution, and show how interpretable ML methods can support tasks such as predicting protein structures, protein function, and protein-protein interactions. This survey also emphasizes the significance of interpreting and visualizing ML-based inference for structure-based protein representations that enhance interpretability and knowledge discovery. Developing such interpretable approaches promises to further accelerate fields including drug development and protein design. 
\end{abstract}





\maketitle

\section{Introduction}\label{sec1}

Recent advances in machine learning (ML) models have transformed the field of protein science, especially structural biology. Proteins are a building block of life, and their function is closely related to their three-dimensional (3D) structure. Indeed, understanding protein structure can often elucidate protein function. However, predicting how a protein folds from its amino acid sequence, which is represented as a list of text characters, has been a major challenge in biology for several decades. Recently, AlphaFold2 (AF2) ~\cite{jumper2021}, a groundbreaking neural network architecture, has made remarkable progress in accurately predicting protein structures that closely resemble experimental data. The rapid and widespread adoption of AF2 by the research community, even those lacking a theoretical background in ML, has been facilitated by user-friendly interfaces like ColabFold~\cite{mirdita2022}. Moreover, interpretable modules in AF2’s neural network architecture, which are directly visualized within protein structures, help researchers interpret the ML model predictions and even apply the system to tasks such as de novo protein design. In particular, the predicted local distance difference test (pLDDT) and predicted aligned error (pAE) assist in screening candidates when designing novel protein binders. These metrics can be utilized for prioritizing designs that are more likely to succeed in experimental validation, supporting increased success rates in laboratory experiments~\cite{bennett2023improving}.

In general, many ML-based models are still like black boxes, so that explaining or interpreting their predictions is often challenging. Therefore, the development of interpretable ML or explainable ML methods, which present the reasoning behind a model’s inference in a manner understandable to humans  \cite{murdoch2019}, is of great importance. Typically, two main approaches are used to explain ML model predictions. The first involves general post-hoc methods applicable to any pre-trained model that clarify the inference process of the model. The second category consists of inherently explainable models designed to predict the final objective metric while providing explanations for their predictions. This group includes decision tree models that offer a set of rules based on input features. Lately, various interpretable methods have been introduced. For instance, in natural language processing, ML architectures based on attention mechanisms use the weight magnitudes in attention layers to interpret predictions, for example identifying sentence parts that influence word translation~\cite{bahdanau2014}. In image processing, architectures based on convolutional neural networks (CNNs) can utilize saliency maps for basic post-hoc explanations of image parts deemed important for predictions \cite{selvaraju2017}. When the input data is represented as a graph, post-hoc methods can provide interpretations of the importance of nodes and edges to the model predictions \cite{ying2019, yuan2020a, yuan2020b, schlichtkrull2020, yuan2021, lin2021, henderson2021}. For example, Miao et al. \cite{miao2022} added a stochastic attention layer to recognize and assign higher weights to important edges in the graph. The development of model-agnostic post-hoc methods has also been an active area of research~\cite{strumbelj2010, ribeiro2016}, with notable advances including the Integrated Gradient (IG) method \cite{sundarajan2017} and Shapley Additive Explanations (SHAP) and its variants \cite{lundberg2017, chen2022}.

Several interpretable methods have been applied to analyze ML-based models focusing on protein-related tasks using structure-based representations. Techniques like Integrated Gradients have been utilized to assign saliency scores to amino acids, aiding in predicting ATP-binding based on a geometric vector perceptron architecture \cite{wang2022}. Similarly, gradient-weighted class activation maps (GradCAM) have been utilized to identify crucial residues for functionality prediction within graph convolutional neural network architectures \cite{gligorijevic2021}. Complementary to the model explanation, the inclusion of metrics trained in a supervised fashion along with the main model objective has added another layer of interpretability to ML models, especially in computational biology. These metrics, visualizable within protein structures, facilitate the identification of amino acids involved in interactions, enriching our understanding of protein complexes, and providing insights that support biological research and development of therapeutics.

In this paper, we explore a variety of structure-based representations for proteins, from the viewpoints of both structural biologists and computer scientists (Section 2), detailing their application in interpretable ML methods for predicting protein structures, functionalities, and understanding protein-protein interactions (Section 3). Our discussion emphasizes the diverse interpretability approaches for ML methods and the creation of visualizations using structure-based representations for broad protein-related tasks (Section 4). We address the challenge of developing new forms of visualization for proteins, posed in~\cite{bourne2022}, that move beyond traditional ribbon diagrams established by Richardson in ~\cite{richardson1981}. Our goal is to enhance the understanding of protein functions, properties, and interactions through designed visualizations and interpretable metrics, thereby advancing current visualization paradigms and simplifying the comprehension of interaction mechanisms in complex scenarios. This effort builds upon the workflow and best practices for interpretability in computational biology introduced by~\cite{chen2022}, with a particular focus on structure-based representations. For insights into interpretability and explainability using sequence-based protein representations, this has recently been covered by \cite{danilevsky2020, ferruz2022}. Overall, our analyses underline paths forward for interpreting and visualizing ML-based inference that promote both model interpretation and biological discovery.
\section{Structure-based protein representations}\label{sec2}

\subsection{Representations in Structural Biology}

\begin{figure}
  \centering
  \subfigure[]{
    \includegraphics[width=0.92\linewidth]{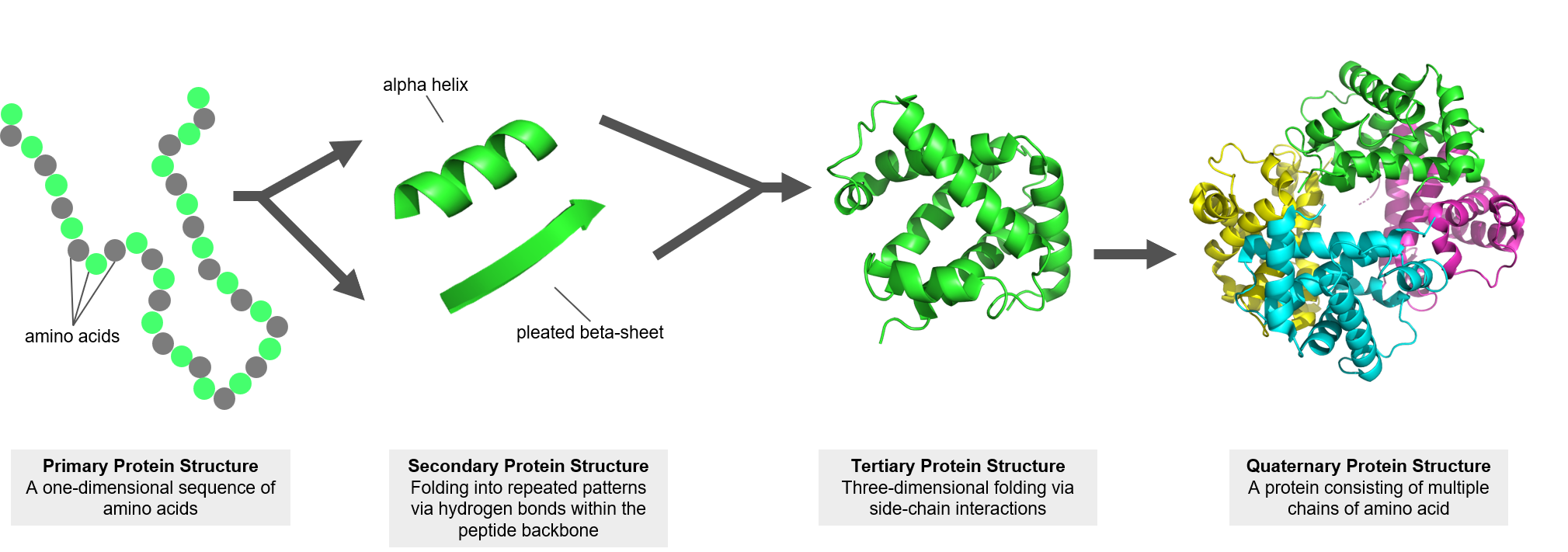}
  }
  \subfigure[]{
    \includegraphics[width=0.23\linewidth]{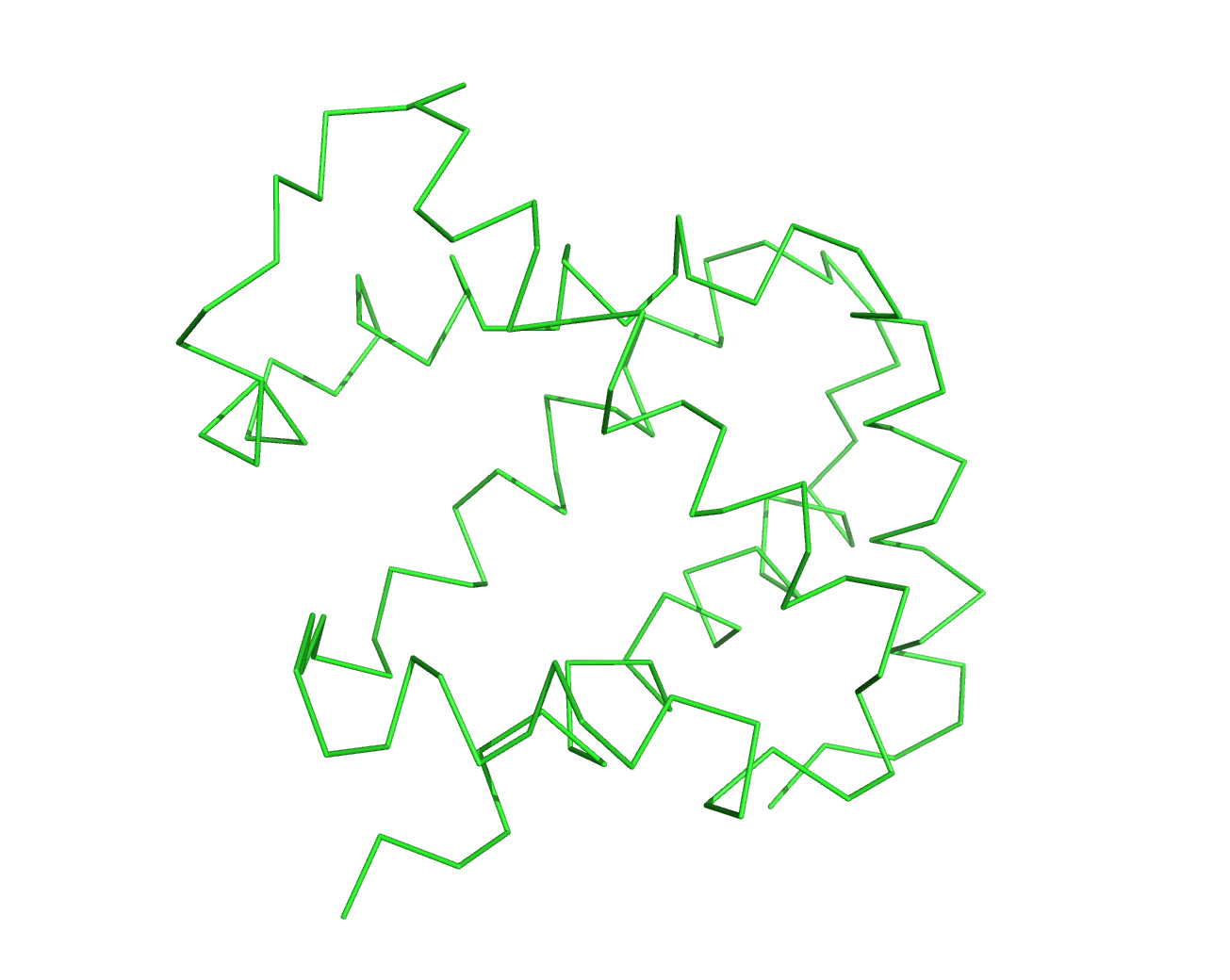}
  }
  \subfigure[]{
    {\includegraphics[width=0.23\linewidth]{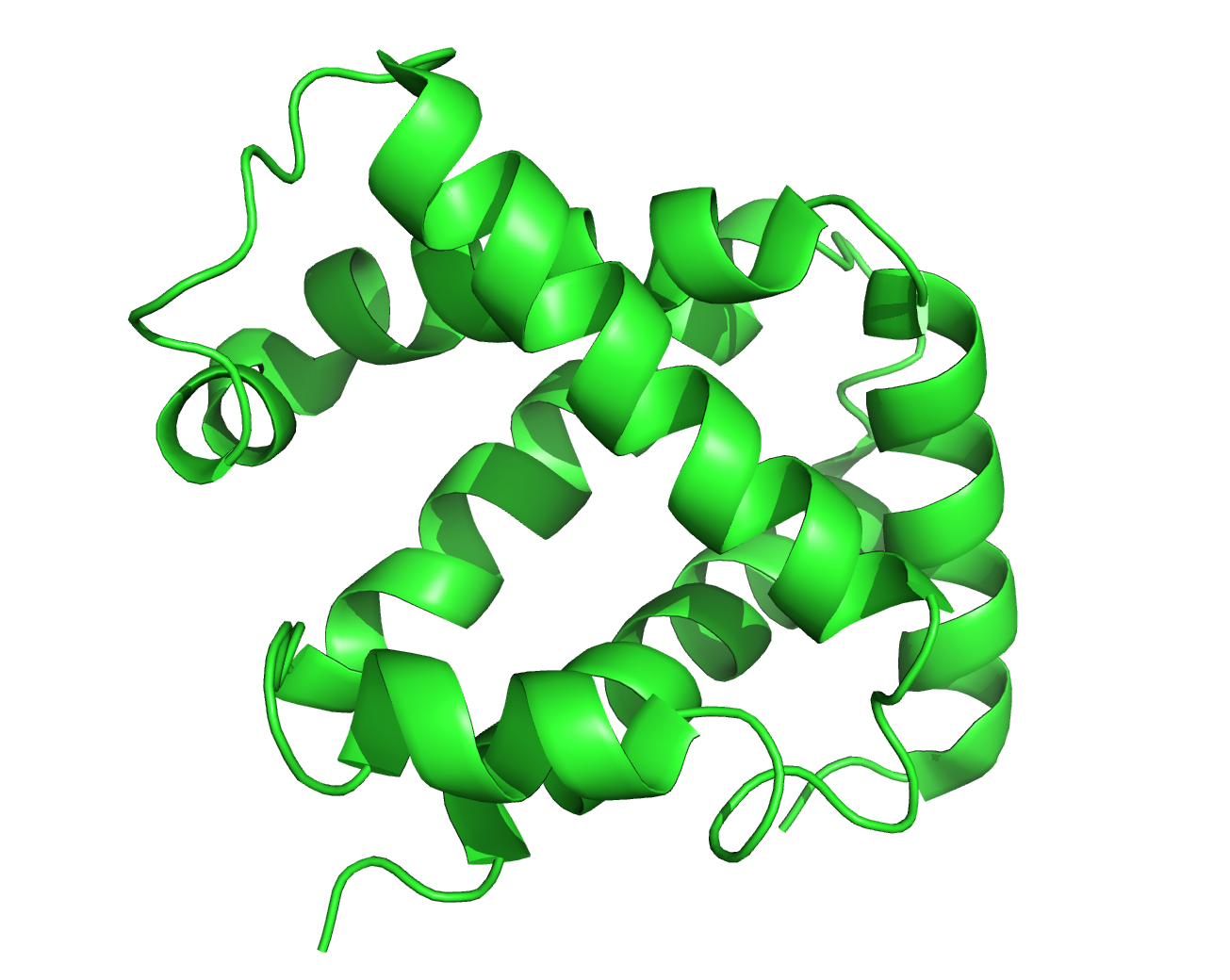}}
  }
  \subfigure[]{
    {\includegraphics[width=0.23\linewidth]{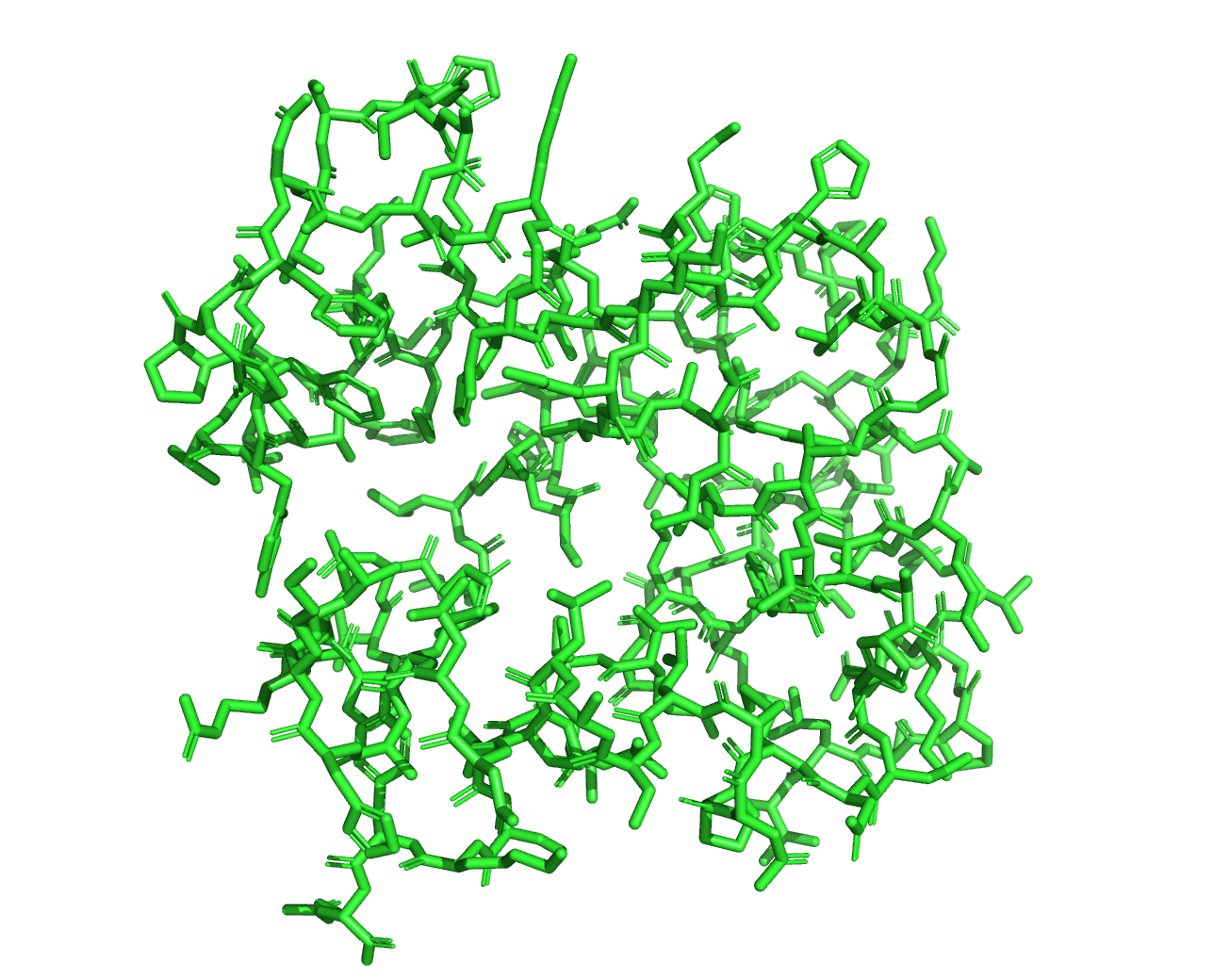}}
  }
  \subfigure[]{
    \includegraphics[width=0.23\linewidth]{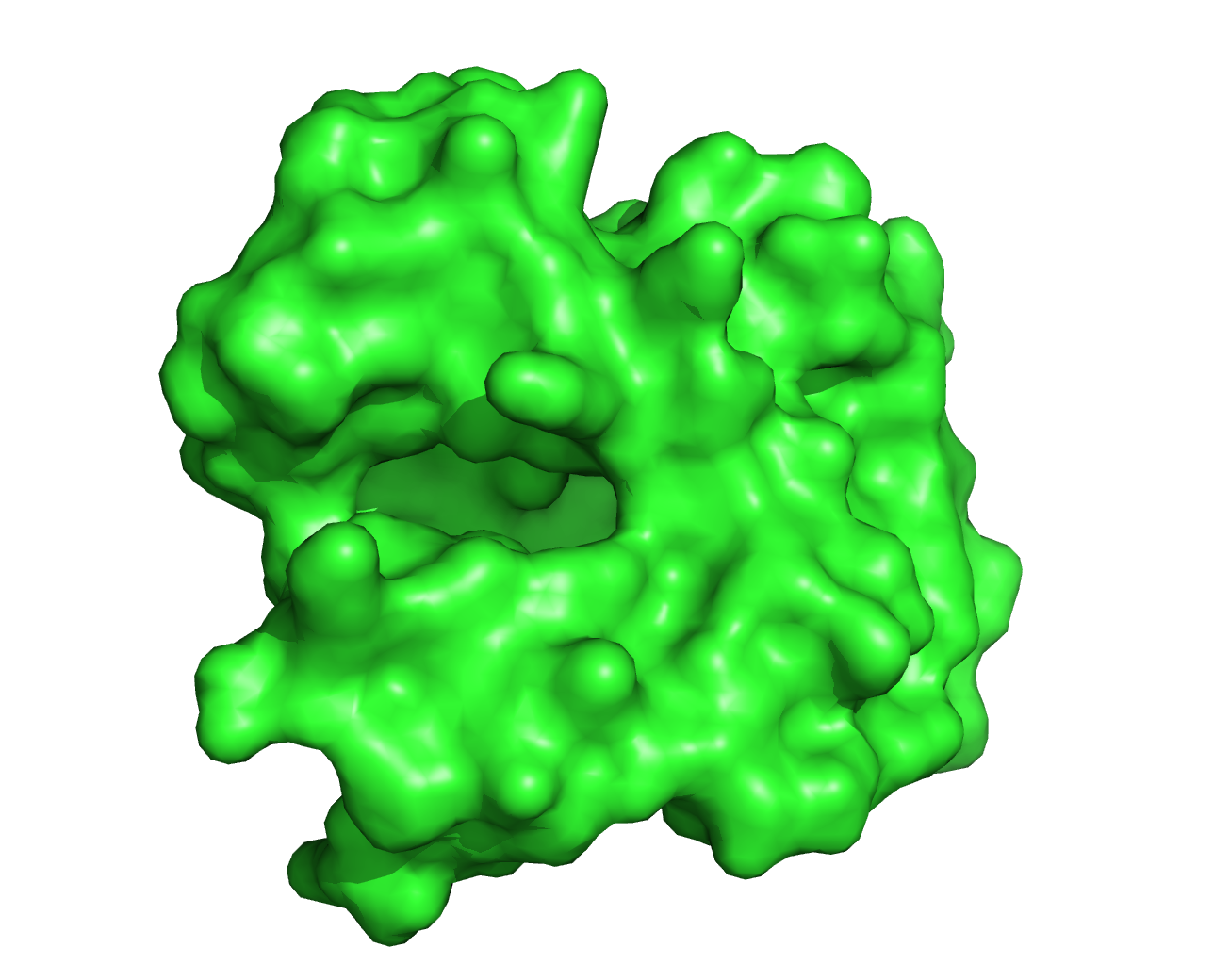}
  }
  \caption{Schematic showing definitions of different levels of protein structure definition and various representations available for structural biologists using PyMol~\cite{pymol}. Figures using the crystal structure of the \textit{human foetal deoxyhaemoglobin} protein (PDB: 1FDH); (a) The \textit{primary} protein structure consists of the sequence of amino acids in the polypeptide chain. The \textit{secondary} protein structure consists of alpha helices and beta sheets formed by hydrogen bonding between atoms in the polypeptide backbone. The \textit{tertiary} protein structure consists of the overall 3-dimensional structure of the folded protein chain. The \textit{quaternary} protein structure consists of the structure formed by multiple interacting amino acid chains; (b) ribbon representation using lines; (c) cartoon representation; (d) all-atom representation using sticks; and (e) surface representation.}
  \label{fig0}
\end{figure}

At the simplest level of protein structure (the primary structure), proteins are conveyed as a sequence of characters, each representing one of the 20 standard amino acids that can be incorporated into a polypeptide chain. Each amino acid carries a specific side chain with defined chemical properties. The polypeptide backbone consists of repeated atoms contributed by each amino acid: a central carbon atom (C), a hydrogen (H) atom, an amino group, consisting of a nitrogen (N) atom and two hydrogen atoms, and a carboxyl group, consisting of a carbon atom, two oxygen (O) atoms, and one hydrogen atom. The sequence of amino acids folds into a stable conformation, the three-dimensional (tertiary) protein structure. A schematic showing a general protein structure is shown in Fig.~\ref{fig0}(a). The atom positions in proteins can be experimentally determined by methods such as Cryo-EM \cite{cheng2015} and X-ray crystallography \cite{bragg1968}. This experimentally derived structural data is then deposited in publicly accessible databases in formats such as PDB and mmCIF \cite{westbrook2003}. These formats contain general information about the protein characterized, experimental parameters, chains, amino acid sequences, and the position of atoms in the structure. The structure in these formats can be visualized computationally using programs such as PyMol \cite{pymol} and ChimeraX \cite{chimerax}.

 Such programs visualize proteins by presenting atomic positions in 3D space. In practice, low-resolution representations of the structure, filtering the positions of atom subsets or coarse-grained representations, are often used for analysis by structural biologists. The ribbon representation is a prominent mode of protein visualization~\cite{richardson1981}that shows the general protein backbone structure (Fig.~\ref{fig0}(b)). This representation can also be seen in Fig.~\ref{fig0}(c) as a ribbon cartoon interpolated over the polypeptide backbone. Ribbon diagrams allow easy visualization of common folding motifs within proteins, including alpha-helices, pleated beta-sheets, and loops. However, as simple and powerful as ribbon diagrams are, visualizing proteins using them is challenging for computer scientists without background knowledge in structural biology. To ease interdisciplinary collaborations, we will present diverse forms of protein structures from both structural biology and machine learning perspectives (presented in Section~\ref{sec3}).

When using high-resolution representations, both backbone and sidechain information are presented. As an example, an all-atom structural representation is shown in Fig.~\ref{fig0}(d). These representations present important information for understanding the interactions between proteins and their stability. In addition, all-atom representations can identify potential structural clashes in the interaction region to facilitate protein design tasks involving structural prediction. High-resolution representations also add information about the surface of the protein, from which geometric and chemical characteristics can be analyzed. A surface representation of a protein is shown in Fig.~\ref{fig0}(e). For ML-based methods, the choice between using low-resolution and high-resolution representations is made based on the tasks being performed and speed-accuracy tradeoffs~\cite{postic2021representations}.
\subsection{Representations in Computational Biology}\label{sec3}

Protein structures can be used in different forms by machine learning models. A common scenario is to use protein structure as a model input. In this case, the ML architecture extracts structural features to predict output labels. When the output label is predicted for a given protein, e.g., whether a protein is functional or not, interpretability is achieved using post-hoc methods that recognize the importance of the input features or using inherent interpretable architectures. If output labels can be associated with protein structures, e.g., predicting coordinate values per residue, interpretability can be achieved by analyzing the predicted labels. In this section, we examine how recently developed ML models handle the specific properties of structure-based representations used as inputs or labels for prediction.

\begin{figure}
  \centering
  \subfigure[]{
    \raisebox{3mm}{
    \includegraphics[width=0.4\linewidth]{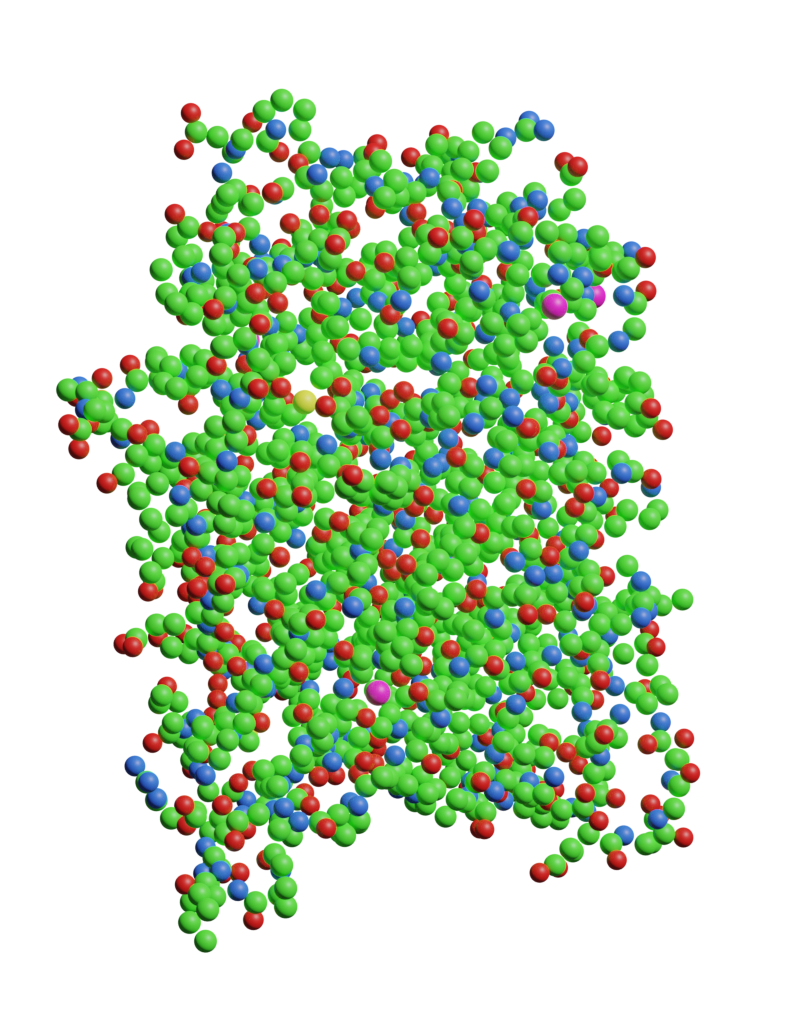}}
  }
  \subfigure[]{
    \raisebox{15mm}{\includegraphics[width=0.4\linewidth]{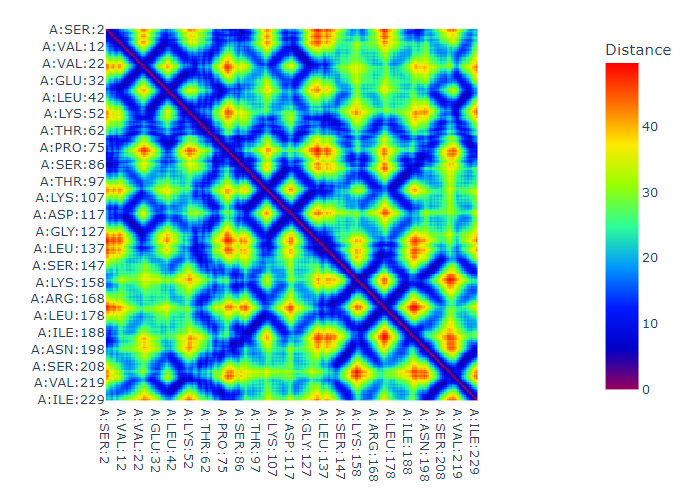}}
  }
  
  \subfigure[]{
    \raisebox{0mm}{\includegraphics[width=0.4\linewidth]{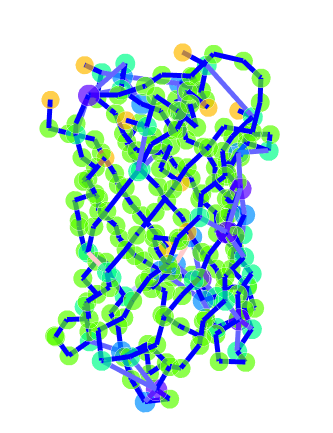}}
  }
  \subfigure[]{
    \raisebox{10mm}{
    \includegraphics[width=0.4\linewidth]{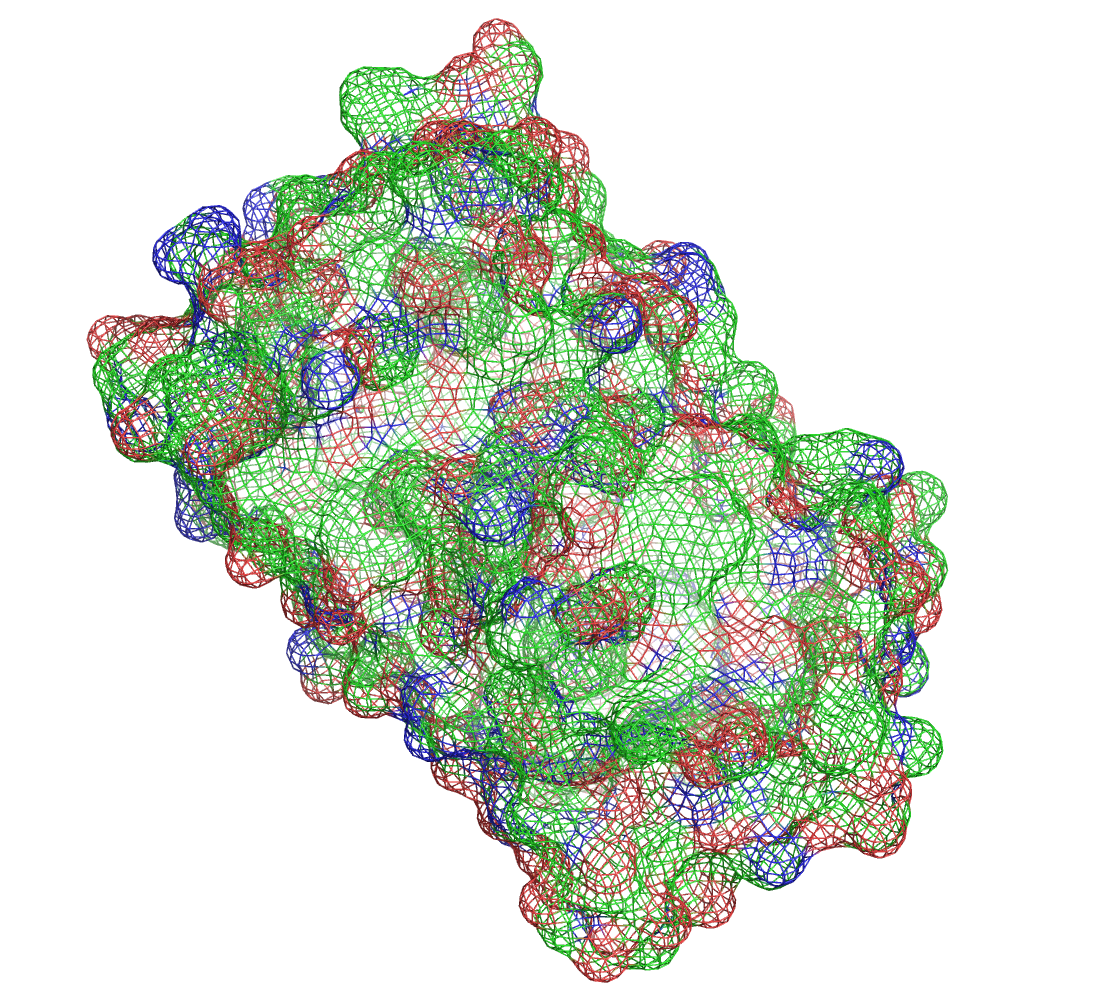}}
  }
  \caption{Examples of structure-based protein representations used in machine learning. Visualization of the green fluorescent protein (PDB: 1ema). (a) Protein structure represented by a point cloud showing backbone atom coordinate positions; (b) Protein structure represented by a distance matrix; (c) Protein structure represented as a graph with residue $C_\alpha$ atom positions represented as nodes and edges defined by neighbors in the amino acid sequence; (d) Mesh representation of the protein surface. Subfigure (a) plotted with Molecular Nodes~\cite{molecularnodes}.  Subfigures (b) and (c) are plotted with Graphein \cite{graphein}. Subfigure (d) plotted with PyMol \cite{pymol}.}
  \label{fig1}
\end{figure}

In its simplest form, a protein can be represented as an atom point cloud (Fig.~\ref{fig1}(a)). However, this representation has inherent problems related to properties that should be fulfilled in protein-related tasks. The following example illustrates these problems. A protein can be located in different parts of the space at different timesteps while having the same conformation. This protein should be mathematically represented as the same entity, even if its structure has different atom coordinates and/or is rotated. However, if we train a traditional ML model only on atomic coordinates for functionality prediction, for example, the model will be sensitive to the coordinate system and current position of the protein, which is undesirable. Therefore, the protein structure representation must be both translation and rotation-invariant. In addition, if the model is generative and the inference is performed in an autoregressive manner, e.g., per-atom or per-residue, it is also necessary for the model to be permutation equivariant. Two methodologies are typically employed by machine learning to account for the above protein properties. The first keeps the use of atom positions in an Euclidean 3D space while learning invariant or equivariant representations leveraging novel methods and model architectures. The second employs the use of internal space representations that are, by definition, invariant, e.g., graphs, distance matrices, and surfaces. These features are pre-calculated and then provided as input for the model. 

For recent approaches that use the structure in Euclidean space as inputs, which include structure prediction networks like AF2 \cite{jumper2021} and RosettaFold (RF) \cite{baek2021, baek2023efficient}, special model architectures are needed. AF2 uses a geometric-aware module named invariant point attention (IPA) that adds 3D points in a local residue frame to the attention formulation to predict values invariant to rotation and translation. In RF, the structure update is based on an SE3-equivariant Transformer architecture introduced in \cite{fuchs2020se}. Recent architectures also include the E(n) Equivariant Graph Neural Networks (EGNN) \cite{satorras2021n} using the 3D structure as the input graph, as presented in reference \cite{chen20233d}. 

As another alternative to using invariant and equivariant architectures, various methods used local frames or proposed canonical poses for proteins while using 3D coordinates as inputs \cite{hiranuma2021improved, martinkus2023abdiffuser}. In DeepAccNet \cite{hiranuma2021improved} voxelization in a local frame is performed for every residue leading to a translation and rotation invariant representation that is then used by a 3D convolutional neural network architecture to predict the accuracy of protein structures. In AbDiffuser \cite{martinkus2023abdiffuser} a frame-averaging methodology is used to obtain the canonical pose of antibodies to train a diffusion-based generative model for antigen-conditioned antibody design. In addition to these properties, the parameterization for the protein structure also affects computational efficiency and accuracy. Various parameterizations \cite{yang2020, jumper2021, baek2023efficient} and coarse-grained representations~\cite{Martini, Primo} have been proposed. The parameterization choice influences the information loss when structure reconstruction is needed and is an important modeling decision in the development of recent diffusion-based structure generation models \cite{FrameDiff, DiffAb, RFdiffusion}. When models predicting structural features in Euclidean space are employed, interpretability is typically achieved by adding per residue metrics that are jointly trained with the main objective by the model \cite{jumper2021, baek2021}.

Invariant internal coordinate representations are also used as input features. These include distance matrices, torsion angles, and graph-based representations. When the pairwise distance between amino acids is presented as a 2-dimensional (2D) image, it is defined as a distance matrix, or distogram if the distance is discretized (Fig.~\ref{fig1}(b)). Distograms can be used by ML-based methods as 2D or 3D images where each pixel is a residue-residue interaction and the channels can be related to the feature being described. Protein distograms hold properties such as symmetry and are usually sparse. Due to this symmetry, distograms contain redundant feature information that can be filtered when used as input for ML models. 

A protein structure can also be represented as a proximity graph over amino acids. In this case, each node represents an amino acid. The edges of the graph represent residue-residue features, e.g., the structural neighborhood of an amino acid (Fig.~\ref{fig1}(c)). A $k$-nearest neighbors approach is often used in which the $k$ amino acids that are closest in distance to a specific node are used to create the graph. Relational reasoning over the graph structure can be performed using ML methods such as Graph Neural Networks (GNNs) \cite{battaglia2018}. When representing a protein as a graph, the features are associated with nodes, i.e., node features, and with edges, i.e., edge features. In its simplest form, the features of a protein graph include a node feature that identifies which amino acid is represented by a specific node, and edge features that depict the distance between two amino acids. The node features and edge features can vary depending on the application and methodology applied. Lately, graph-based protein representations have been highly effective in tackling protein sequence design as demonstrated by \cite{ingraham2019, dauparas2022}. For models using internal representations, interpretable metrics are normally associated per-node, e.g., for each residue, or per-edge, e.g., residue-residue interactions.

Interactions between proteins are related to the geometric and chemical characteristics of the surface region in which the interaction occurs. To encode these features, ML methods like MaSIF (molecular surface interaction fingerprinting), dMaSIF, and DeepSurf \cite{gainza2020, igashov2022, mylonas2021} make use of a molecular surface representation to describe the protein structure (Fig.~\ref{fig1}(d)). These methods assume that invariant structure-based presentations related to protein surfaces are better indicators of how proteins interact. Using a surface-based representation, MaSIF \cite{gainza2020} uses geodesic convolutional layers to learn surface features that are used to predict protein-protein interactions and functions. Meanwhile, dMaSIF \cite{igashov2022}] represents the protein molecular surface using a point cloud in which each point has associated features that can be computed in an end-to-end manner. Since dMaSIF \cite{igashov2022} does not need to calculate the mesh and the patches in a pre-processing step, the method is computationally efficient while achieving the same accuracy as MaSIF \cite{gainza2020}. Implicit neural representations have also been explored to model protein surfaces by Sun et al \cite{sun2024dsr}. Following an approach similar to geodesic convolutional neural networks (GCNNs) \cite{masci2015}, Wang et al \cite{wang2023learning} learn molecular surface representations on a 2D Riemannian manifold instead of in the 3D Euclidean space. The HMR framework proposed in \cite{wang2023learning} offers multi-resolution surface features while the learned representation is rotation and translation invariant. Metrics learned by these methods can be used to assign parts of the surface associated with different functionalities and interactions.
\section{Interpretable Machine Learning for Protein Structural Biology}\label{sec4}

In this section, we showcase interpretable ML applications in structural biology by discussing examples proposed for three protein-related tasks: structure prediction, functionality prediction, and protein-protein interactions. We focus on ML-based methods that allow the creation of interpretable visualizations utilizing the structure-based protein representations presented in Section~\ref{sec3} (Table~\ref{tab:survey}). The four ML methods surveyed are (i) a method for protein structure prediction based on neural network blocks called evoformers; (ii) a method based on gradient boosting decision trees that take as input a vector with features representing residue-residue interactions; (iii) a method based on graph convolutional neural networks that calculate embeddings per-residue; and (iv) a method based on geodesic convolutional networks that utilize a protein surface representation.

\begin{table}
  \caption{Survey of interpretable methods using structure-based protein representations.}
  \label{tab:survey}
  \centering
\begin{NiceTabular}{ |p{2.8cm}|p{1.0cm}|p{4.3cm}|p{5.0cm}| }[hvlines]
 \multicolumn{4}{l}{\textbf{(1) Functionality Prediction}} \\
 \hline
 Name & Inputs & Interpretability Type & Description \\
 \hline
 LM-GVP \cite{wang2022} & \Block{4-1}{Seq/Str} & Post-hoc using IG~\cite{sundarajan2017} applied to latent representation & Combine features from a protein LM to structural node embeddings of a GNN \\
 DeepFRI \cite{gligorijevic2021} &   & \Block{2-1}{Post-hoc using GradCAM~\cite{selvaraju2017} applied to latent representation} & Use features from an LM and structural features in a GCN architecture \\
 HEAL \cite{gu2023hierarchical} & & & Hierarchical GCN trained with contrastive learning \\
 Graph-GradCAM \cite{ravichandran2023predicting} & & Post-hoc using a proposed Graph GradCAM method with GNN & Interpretable GNN architecture combining sequence and structural features \\
 \hline
 Lee et al. \cite{lee2022structure} & Str & Paths learned by a decision-tree based method & Interpretable method based on gradient boosting decision trees \cite{lightgbm} \\
 \hline
 \multicolumn{4}{l}{\textbf{(2) Structure Prediction}} \\
 \hline
 Name & Inputs & Interpretability Type & Description \\
 \hline
 ESMFold \cite{ESMFold} & \Block{2-1}{Seq} & \Block{5-1}{Confidence metrics per residue and per residue pairs} & End-to-end single-sequence architecture using a protein LM and equivariant transformers \\
 OmegaFold \cite{OmegaFold} & & & End-to-end single-sequence architecture using a protein LM and geoformer layers \\
 AF2 \cite{jumper2021} & \Block{4-1}{Seq/Str} &  & Evoformer architecture and a structure module \\
 RF \cite{baek2021, baek2023efficient} & & & Three-track network using attention modules and SE(3)-Transformers \\
 OpenFold \cite{OpenFold} & & & Evoformer architecture~\cite{jumper2021} with optimized training \\
 IgFold \cite{ruffolo2022} & & Confidence metrics per residue & Architecture using an antibody LM, graph transformers, and IPA modules \\
 \hline
 \multicolumn{4}{l}{\textbf{(3) Protein-Protein Interactions}} \\
 \hline
 Name & Inputs & Interpretability Type & Description \\
 \hline
 HIGH-PPI \cite{gao2023hierarchical} &\Block{2-1}{Seq/Str} & GNNExplainer~\cite{ying2019} to subgraph representations & Hierarchical GCN for protein structures and network \\
 ScanNet \cite{tubiana2022} & & Labels per residue & Interpretable geometric DL architecture using spatio-chemical features \\
 MaSIF \cite{gainza2020} &\Block{4-1}{Str/Sur} & \Block{3-1}{Labels associated with surface patches} & Geometric DL architecture using geometric and chemical features \\
 dMaSIF \cite{sverrisson2021fast} & & & End-to-end differentiable geometric DL architecture \\
 DeepSurf \cite{mylonas2021} & & & LDC-ResNet architecture applied to protein surfaces \\
 HMR \cite{wang2023learning} & & Labels associated with a surface manifold & Manifold harmonic message passing using surface representations \\
 PeSTO \cite{krapp2023pesto} & Str & Labels per residue & Geometric transformer architecture using protein structures \\
 \hline
\end{NiceTabular}
\end{table}

\subsection{Protein structure prediction}
\label{subsec3-3}

When using structure prediction networks such as AF2, RF, and IgFold \cite{jumper2021, baek2021, ruffolo2022} for protein design, a crucial aspect of these models is their capacity to both predict structures accurately and to provide confidence metrics for the structural predictions. These metrics provide crucial insights for designers, particularly in identifying unstable regions of the protein. As discussed in Section~\ref{sec1}, the two primary metrics provided by AF2 are the pLDDT and pAE. The pLDDT metric is predicted per residue and is typically used to color the protein structure, indicating the quality of the AF2 prediction. The pAE metric is calculated for residue-residue interactions and is usually plotted as a 2D image resembling a distogram or contact map. The pAE metric is usually used to filter candidates in the design of de novo protein complexes \cite{bennett2023improving}. 

\begin{figure}[t]
 \subfigure[]{
  \raisebox{0mm}{\includegraphics[width=0.3\linewidth]{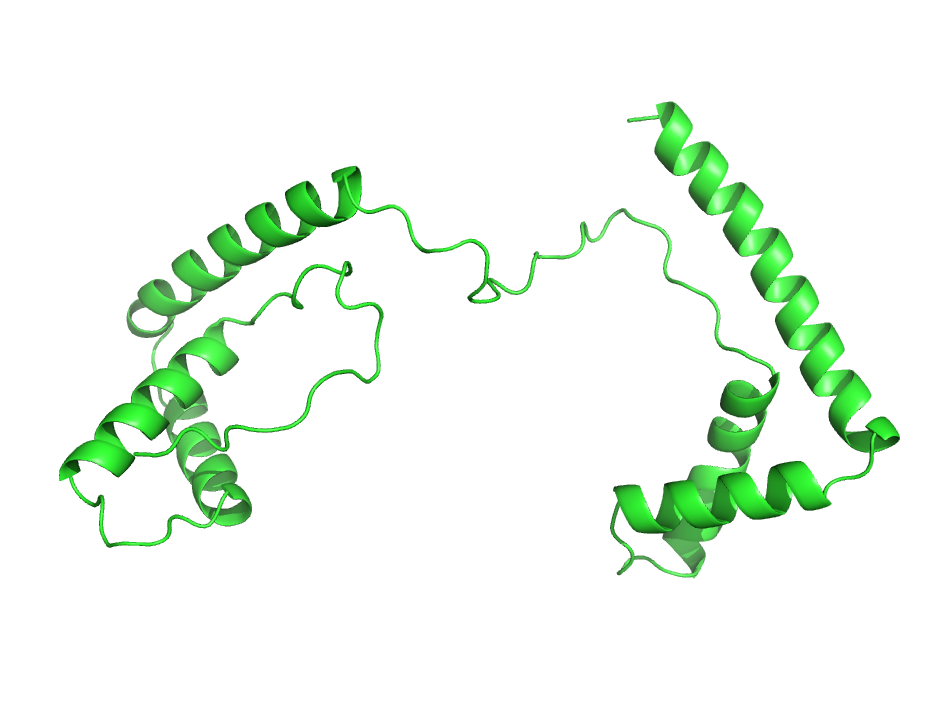}}
 }
 \subfigure[]{
  \raisebox{0mm}{\includegraphics[width=0.3\linewidth]{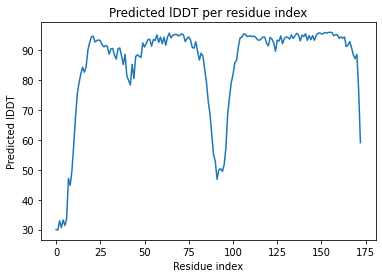}}
 }
 \subfigure[]{
  \raisebox{0mm}{\includegraphics[width=0.3\linewidth]{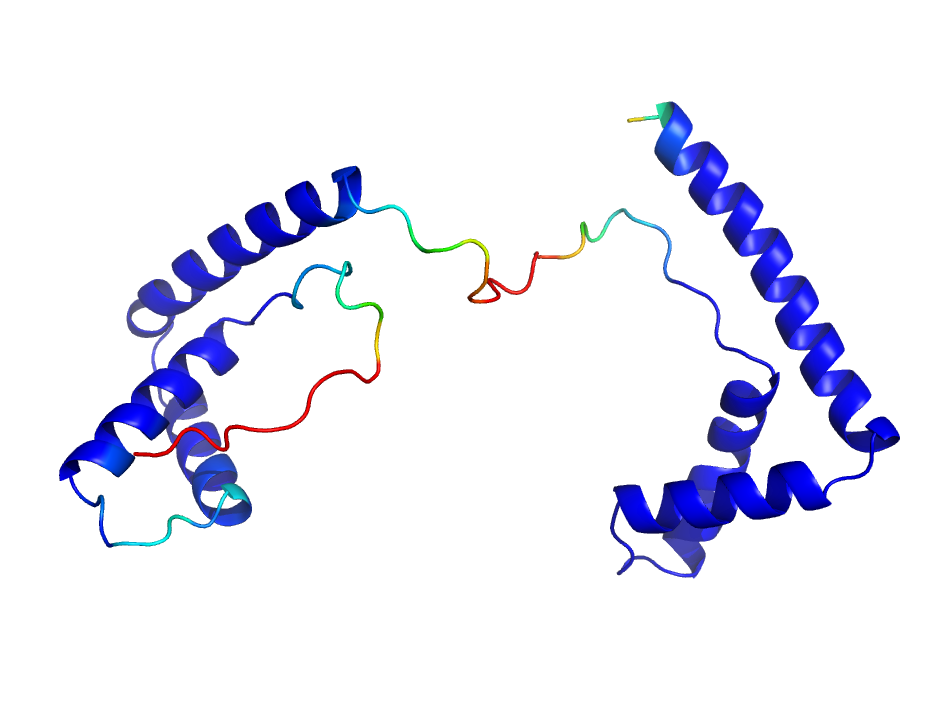}}
 }
  \caption{Structure predicted by ColabFold \cite{mirdita2022} for a tandem HMG box domain from the HMGB1 protein (PDB: 2YRQ) (a) structure colored by chain; (b) 2-dimensional plot of the pLDDT metric by residue index; and (c) structure colored by pLDDT score. Blue residues represent amino acids in which AF2 has high confidence in the prediction. Red residues represent amino acids in which AF2 has low confidence in the prediction.}
 \label{fig6}
\end{figure}

In this section, we applied AF2, as implemented in ColabFold \cite{mirdita2022}, to predict the structure of a protein (PDB: 2YRQ). The predicted structure, shown in  Fig.~\ref{fig6}(a), is colored uniformly, thus lacking visualization of a confidence score for predicting different regions. A protein designer would thus need to rely on heuristic methods to evaluate the structures. In Fig.~\ref{fig6}(b), a 2D plot showing the pLDDT score by residue index is shown. The valleys convey where the AF2 model is not confident, but associating the residue index with the protein structure remains challenging. Finally, in Fig.~\ref{fig6} the pLDDT values are shown in the structure using a color-based plot, i.e., blue regions represent high confidence while red regions represent low confidence in the prediction. Parts of the structure with low confidence are clearly visible and can help the evaluation of model predictions. This information can be used, for example, to identify parts of the protein to target for optimization.

\begin{figure}[t]%
 \subfigure[]{
  \raisebox{3mm}{\includegraphics[width=0.3\linewidth]{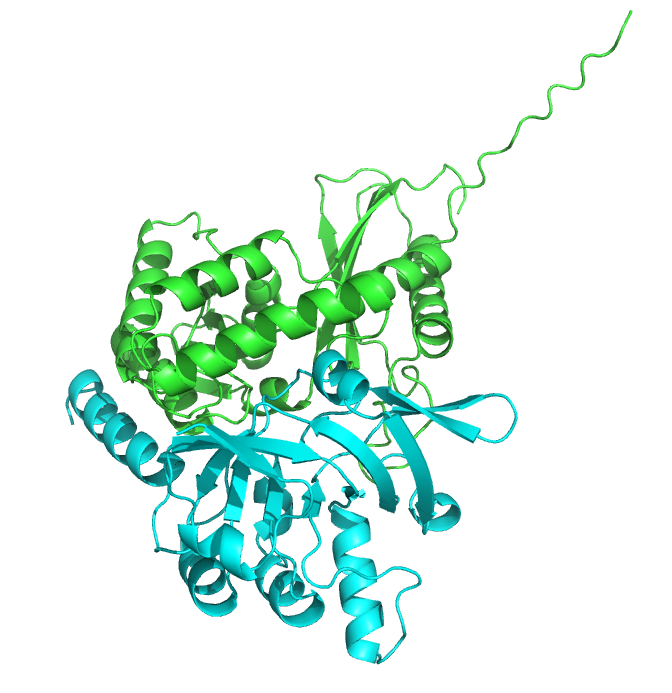}}
 }
 \subfigure[]{
  \raisebox{0mm}{\includegraphics[width=0.3\linewidth]{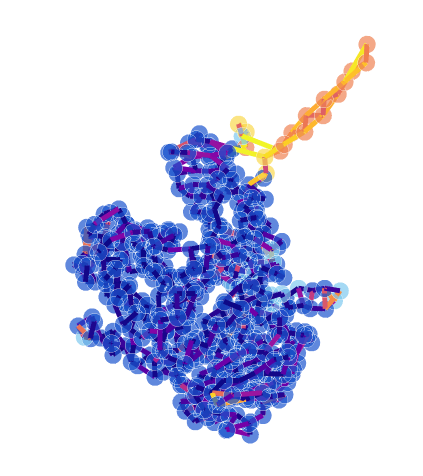}}
 }
 \centering
  \subfigure[]{
  \raisebox{0mm}{\includegraphics[width=0.3\linewidth]{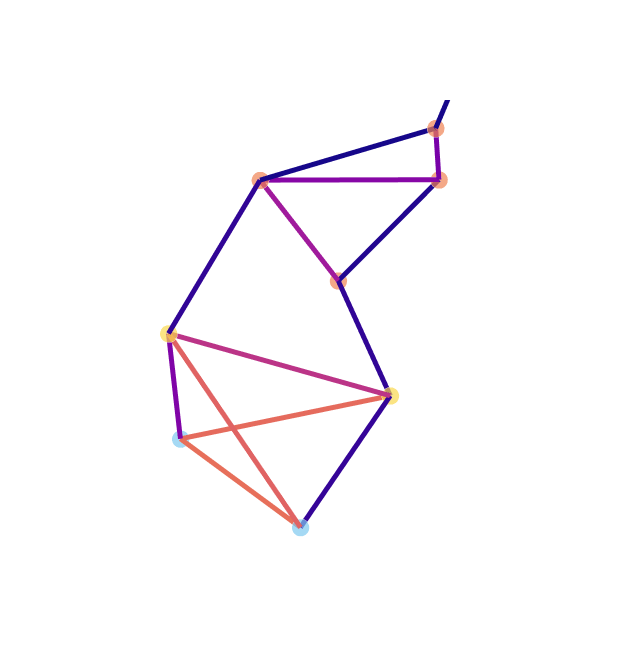}}
 }
\caption{Graph-based visualization of confidence metrics by AF2-Multimer \cite{evans2021} using ColabFold \cite{mirdita2022} predicted for a protein (PDB: 7XKY, Fumarate hydratase apo-protein complex) (a) Predicted structure by AF2 visualized using PyMol \cite{pymol}; (b) Predicted structure by AF2 visualized as a graph using Graphein \cite{graphein}. Nodes represent residues in 3D space. Edges represent the $k$-nearest neighbors of a residue, in which $k$ is set to 3. Nodes are colored by pLDDT in which lighter color means higher predicted error on the atom position; (c) Zoomed in view to show the internal structure of the plot presented in (b) in which edges shown in a lighter value are predicted to have higher pAE.}
\label{fig4}
\end{figure}

We also applied a graph-based visualization tool~\cite{graphein} to analyze the pLDDT and pAE metrics by AF2. The resulting predicted structure is visualized as a 3D graph, with pLDDT values used to color nodes and pAE values used to color edges that represent residue pairs. The structure predicted by the AF2-Multimer model for the protein sequence of the Fumarate hydratase apo-protein complex (PDB: 7XKY) is visualized in Fig.~\ref{fig4}. The structure is represented in a cartoon-like form in Fig.~\ref{fig4}(a), while its corresponding graph-based representation is shown in Fig.~\ref{fig4}(b). In Fig.~\ref{fig4}(b), each node in the graph denotes a residue in 3D space, and k-nearest neighbors are used to define three neighbors for each residue to aid the visualization of edges. Alternatively, other methods such as a distance threshold or chemical interactions can also be used to define edges \cite{graphein}. The pLDDT values are used to color the nodes in Fig.~\ref{fig4}(b). A graph-based approach is also useful for visualizing residue-residue interactions, as shown in Fig.~\ref{fig4}(c)) where a magnified view of the internal structure is shown, and the edges are colored by pAE. The lighter-colored residue pairs indicate a higher predicted aligned error for that interaction.

Recent structure prediction networks incorporate interpretable modules through the use of confidence metrics. Such metrics calculated per residue can be effective in highlighting the residues that are important in the protein structure. Conversely, for metrics related to residue-residue interactions or higher-order interactions, the visualization can be complex. When patterns are not evident from a 2D representation, alternative visualizations become necessary. For example, a recent tool for pAE visualization has been proposed in \cite{elfmann2023pae}. Figure~\ref{fig4} demonstrates how graph-based visualizations provide one possible alternative for analyzing the characteristics of the internal parts of the structure. 

\subsection{Protein functionality prediction}
\label{subsec3-1}

Optimizing the functionality of a protein is important for many industries that are developing agricultural and therapeutic applications \cite{huang2016coming}. For these purposes, training models to predict protein functions based on sequence and/or structure is crucial. For example, these models can act as feedback to optimize algorithms \cite{stanton2022accelerating, kirjner2023optimizing} or identify regions in the structure that underlie specific protein functions. We survey two ML-based methods and how their interpretable modules can be important for protein engineering.

\begin{figure}[t]%
\centering
\includegraphics[width=0.72\textwidth]{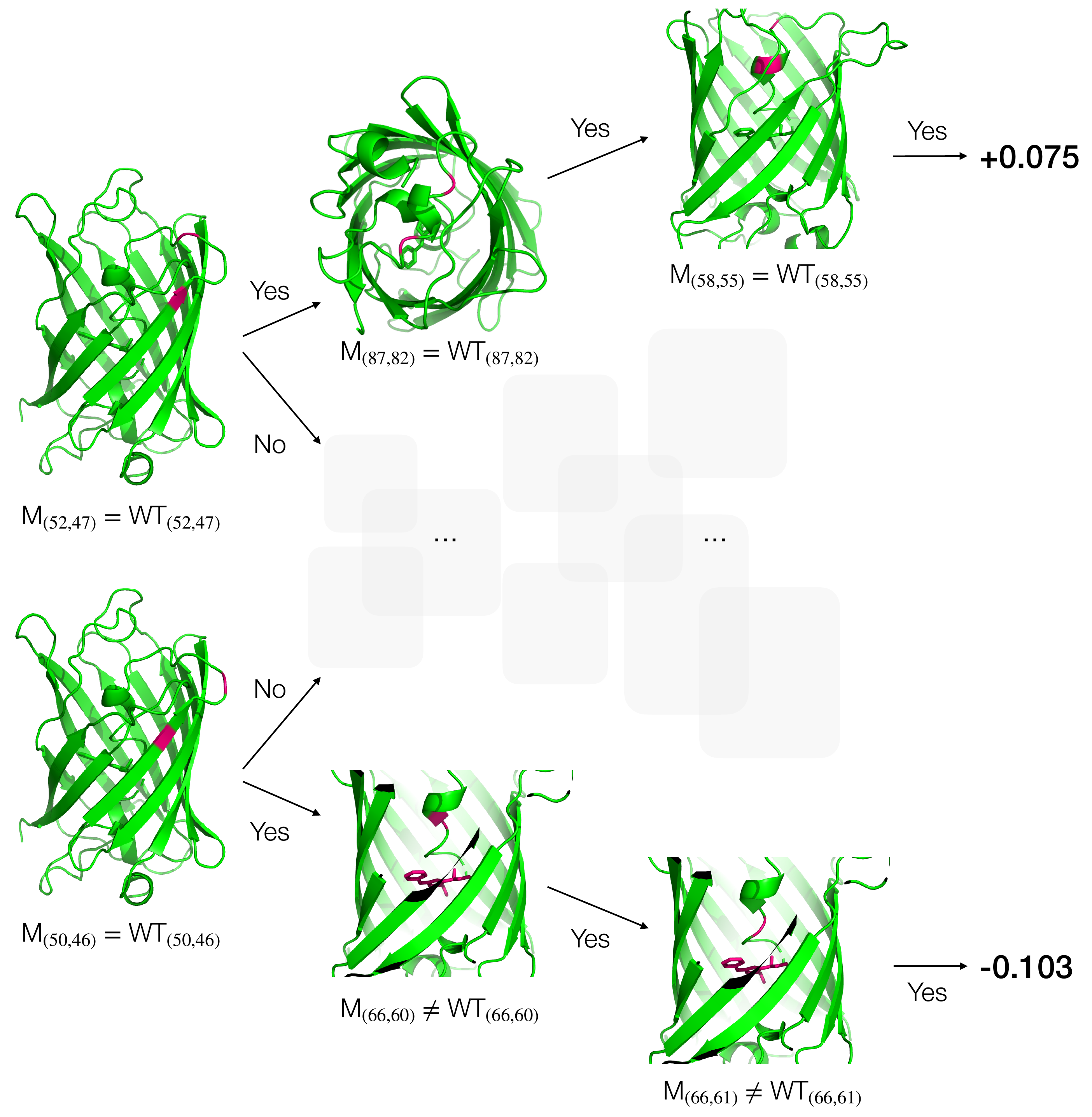}
\caption{Visualization of the nodes leading to the largest increase and the largest decrease of the log-fluorescence value in a GBDT-based predictor. The top part of the image shows 3 nodes of the subtree that lead to the largest increase in log-fluorescence value. The bottom part of the image shows 3 nodes of the subtree that lead to the largest decrease in log-fluorescence value. Each node demonstrates the decision criteria and the positions of the amino acid pair. The amino acid pair related to the input feature being analyzed is highlighted in red. The visualization of the interaction is created using PyMol \cite{delano2002pymol} with the cartoon representation of the wild-type protein (PDB: 1EMA). The terms $\text{M}_{(i,j)}$ and $\text{WT}_{(i, j)}$ represent the distogram value for the interaction between the i-th and j-th residue in the protein sequence for the mutant structure and wild-type structure, respectively.}
\label{fig2}
\end{figure}

\subsubsection{Predictions based on Paths in Decision Trees}

First, we showcase the interpretability of paths learned by an ML method based on decision trees trained for functionality prediction. For this, we employ a distogram, akin to the one depicted in Fig.~\ref{fig1}(b), to filter significant residue-residue interactions, and create a vector, which is utilized as the model's input. This input vector denotes the predicted structure of a mutant protein. Given this input vector, the model tries to predict the functionality value of a given protein. The objective is to predict the log-fluorescence value of the protein. In order to achieve this, we used a Gradient Boosting Decision Trees (GBDT) method \cite{lightgbm} for training the model on the \emph{Aequorea victoria} (avGFP) green fluorescent protein dataset published by Sarkisyan \emph{et al} \cite{sarkisyan2016local}. In GBDT, the criteria of each node are learned to minimize the predictive error, and the values of leaf nodes are summed to obtain the final predicted value. Lately, extensive research has investigated and improved the performance of GBDTs \cite{xgboost, lightgbm, catboost} given their inherent interpretability.

With the trained model, we show how each residue-residue interaction affects the final prediction for a mutant protein. In Fig.~\ref{fig2} two paths are shown, one with the decisions leading to the largest value increase and one with the decisions leading to the largest value decrease for the log-fluorescence value. The decisions leading to the largest increase are interpreted as the ones that are likely to keep the mutant protein with the desired functionality, while the decisions leading to the largest decrease are interpreted as the ones most likely to make the mutant protein lose the desired functionality. It is observed in Fig.~\ref{fig2} that, for the decision tree learned by the model analyzed, the path leading to the largest decrease in log-fluorescence value includes the chromophore region, which is closely related to function, i.e., fluorescence of the avGFP protein~\cite{sarkisyan2016local}. This example showcases the use of inherent interpretable ML methods to discover important interactions in functional regions of the protein structure.

\subsubsection{Importance per-residue in Graph Convolutional Neural Networks predictions}

\begin{figure}[t]%
  \subfigure[]{
    \raisebox{0mm}{\includegraphics[width=0.40\linewidth]{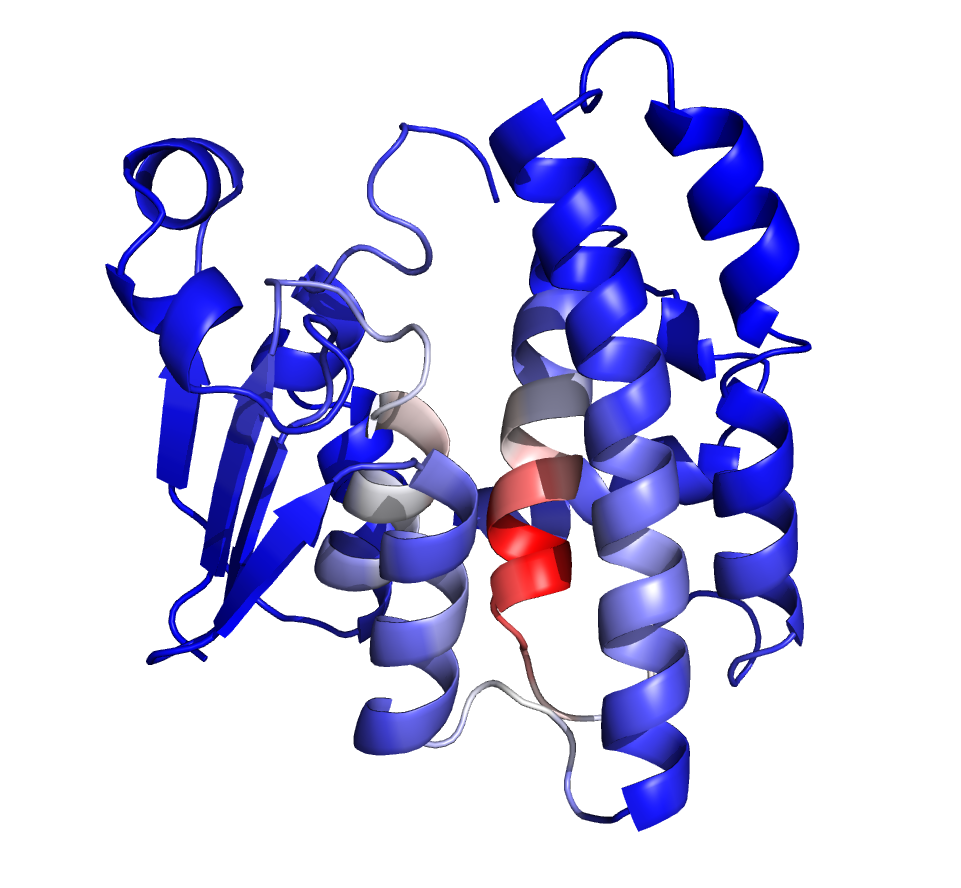}}
  }
  \subfigure[]{
    \raisebox{3mm}{\includegraphics[width=0.40\linewidth]{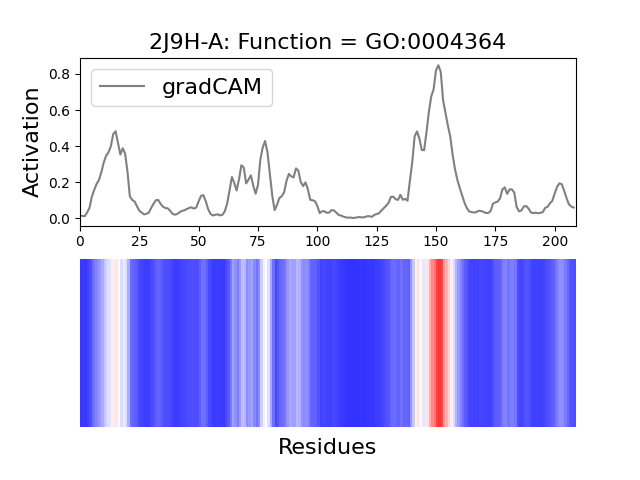}}
  }
\caption{Per-residue calculated by gradCAM \cite{selvaraju2017} using DeepFRI \cite{gligorijevic2021} model (a) Per-residue importance shown in PyMol \cite{pymol} for part of the glutathione transferase P1-1  protein (PDB: 2J9H-A); (b) Per-residue importance shown as (top) the continuous activation values identified by gradCAM activity profile and (bottom) per-residue importance shown as a heatmap. More salient values are in red while less salient values are colored in blue. DeepFRI predicts the protein (PDB: 2J9H-A) as having the glutathione transferase activity (GO:0004364).}
\label{fig3}
\end{figure}

We also present the interpretable characteristics of a functionality prediction method called DeepFRI \cite{gligorijevic2021} based on graph convolutional neural networks (GCNs). DeepFRI uses both sequence and structure representations to learn per-residue embeddings, which are then used by a simple multi-layer perceptron (MLP) network for prediction. The sequence representation is created using embeddings from a protein language model, while the structure representation is modeled as a contact map that utilizes a distance threshold between residues. The model aims to predict protein functions, where the classification schemes used are those proposed by the Gene Ontology (GO) Consortium \cite{geneontology} and the Enzyme Commission (EC) numbers \cite{enzymedatabase}. By definition, the DeepFRI model \cite{gligorijevic2021} is not inherently interpretable. To identify residues important for function prediction, interpretability is incorporated into the trained model by using a gradient-weighted class activation map (gradCAM) \cite{selvaraju2017}. GradCAM, a post-hoc method, is used to identify parts of the input for a given prediction. In DeepFRI, gradCAM is applied directly to the per-residue embeddings learned by the GCNs.

Using the trained DeepFRI model described in \cite{gligorijevic2021}, the importance per-residue assigned by gradCAM is depicted in Fig.~\ref{fig3}. The results in Fig.~\ref{fig3} illustrate the GO term prediction of human glutathione-S-transferase (PDB: 2J9H-A). The gradCAM output is presented in the top row of Fig.~\ref{fig3}(b) and shows higher values for residues that are more significant for the prediction. The heatmap of these values is displayed in the bottom row of Fig.~\ref{fig3}(b), showing distinct regions that are responsible for predicting the target protein’s glutathione transferase activity (GO:0004364). Although the peaks appear distant along the protein sequence in Fig.~\ref{fig3}(b), a structural analysis, shown in Fig.~\ref{fig3}(a), reveals that they are located proximally at the structural level. Overall, even though no information about co-factors, active sites, or site-specificity is given during training, the use of GCNs combined with gradCAM identifies key segments of the structure essential for functionality prediction.

In DeepFRI the final embeddings per-residue account for the embeddings of neighboring residues, and the final heatmap values used to highlight residues in Fig.~\ref{fig3} are calculated using averaged windows. Therefore, the interpretation of these plots is related to the importance of the residue and its neighboring residues in the final prediction. The approach shown in this section is closely related to the per-residue embeddings calculated by recent protein language models \cite{rives-lm, ferruz-lm, nijkamp-lm} proposed in the literature, which has also been applied for functionality prediction. Hence, post-hoc methods for interpretability can be applied similarly to protein language models. For language models based on recent architectures such as Transformers \cite{vaswani2017} and BERT~\cite{devlin2018bert}, another possibility is to use the output of attention layers, which are inherent interpretable modules, to visualize the importance of each residue for the final prediction. Multimodal approaches combining protein sequence/structure representations with biomedical data, as described in \cite{xu2023protst} present possibilities for analysis that give counterfactual explanations via natural language generation.

\subsection{Understanding protein-protein interactions}
\label{subsec3-4}

In this section, we present interpretable metrics given by a surface-based method to predict protein-protein interactions (PPI) using geometric deep learning. Given the complexity of analyzing PPI based on atom positions and interactions between residues, MaSIF \cite{gainza2020} hypothesizes that analyzing PPI using a surface representation of the protein structure could be effective for learning common fingerprints important for interaction. MaSIF learns optimized embeddings for patches on the protein surface that incorporate geometric and chemical features. These embeddings can be used for different downstream tasks, such as protein pocket-ligand prediction and protein-protein interaction site prediction. For site prediction, MaSIF produces a surface score that reflects the likelihood of each patch participating in interactions. Recently, these scores have been proven to be effective in searching protein-protein matches on de novo designed proteins \cite{gainza2023novo}. In \cite{gainza2023novo}, MaSIF predictions are used to define possible surface targets for interactions and to search whether two protein surfaces are likely to interact.

\begin{figure}[t]%
 \subfigure[]{
  \raisebox{5mm}{\includegraphics[width=0.45\linewidth]{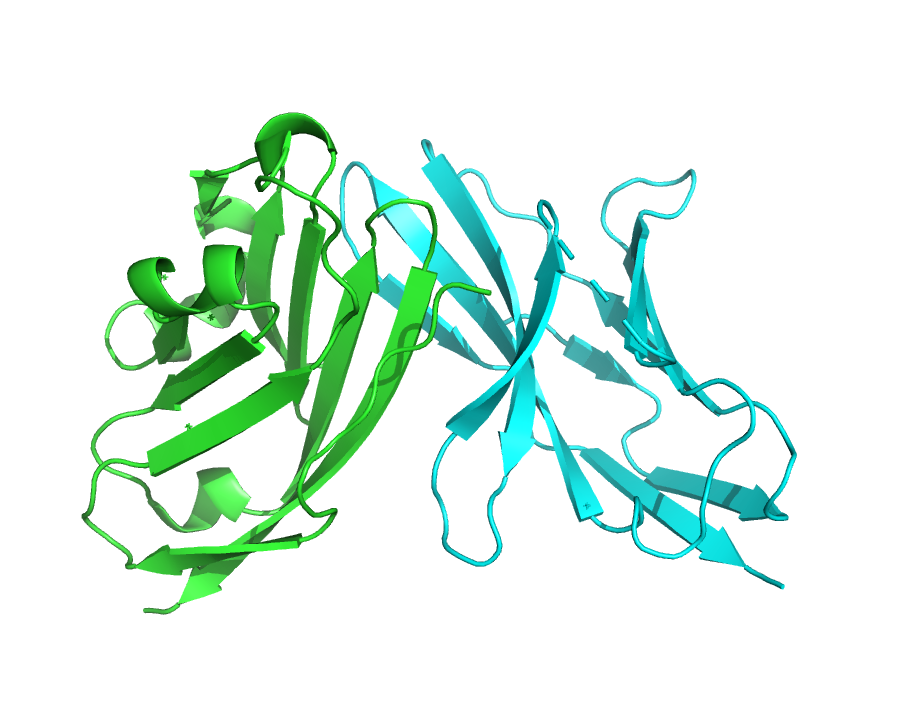}}
 }
 \subfigure[]{
  \raisebox{5mm}{\includegraphics[width=0.45\linewidth]{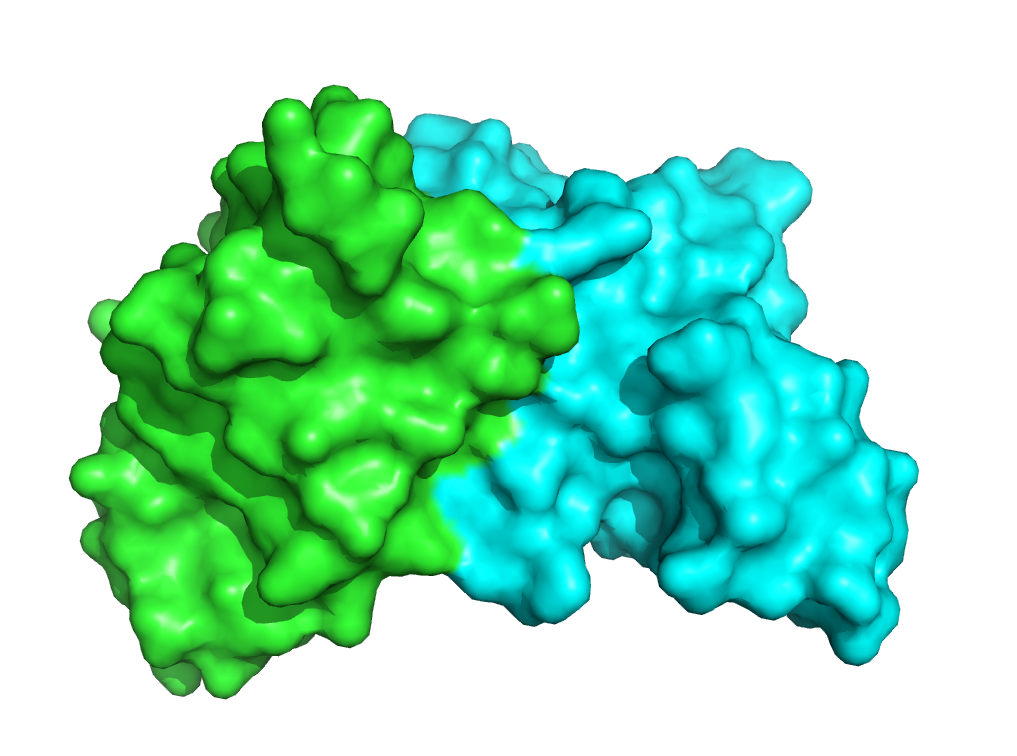}}
 }
  
 \subfigure[]{
  \raisebox{5mm}{\includegraphics[width=0.45\linewidth]{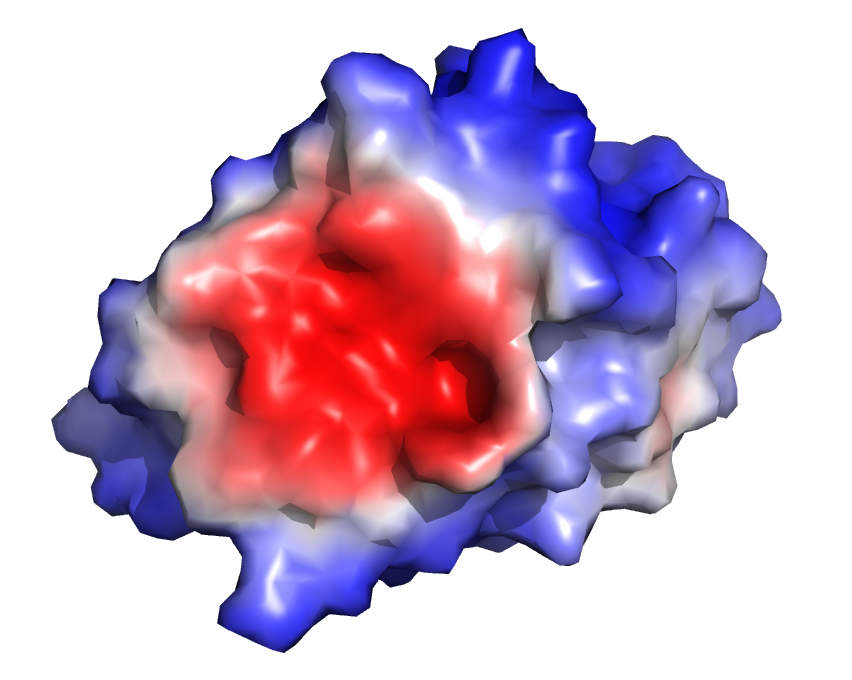}}
 }
 \subfigure[]{
  \raisebox{5mm}{\includegraphics[width=0.45\linewidth]{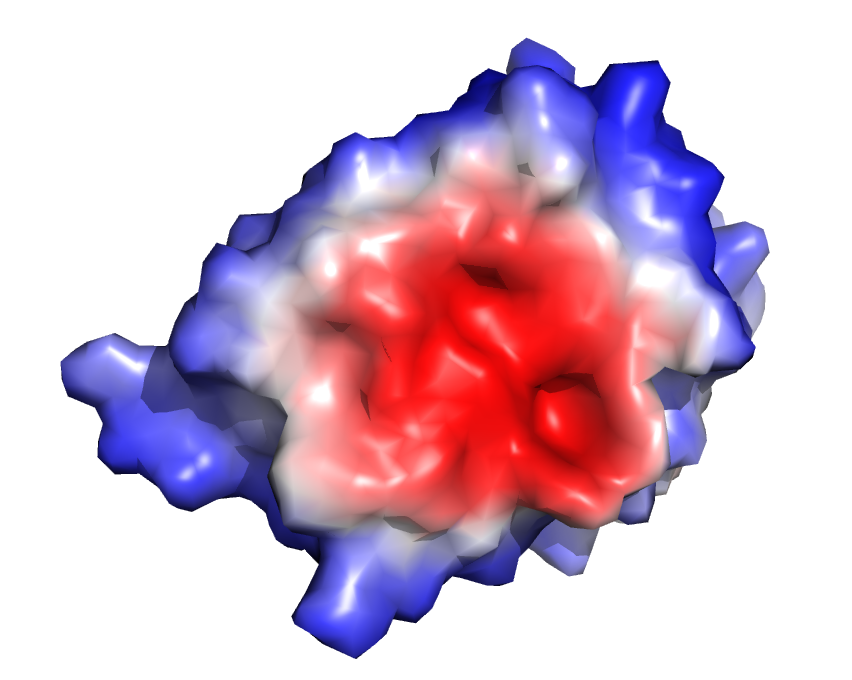}}
 }
\caption{Surface protein-protein interaction site predictions using MaSIF \cite{gainza2020} (a) Structure of the ligand-bound human programmed death-1 (PD-1) protein (PDB: 4ZQK) in cartoon representation (b) Protein surface (PDB: 4ZQK) (c) Interaction fingerprints in chain A (PDB: 4ZQK\_A) calculated by MaSIF. (d) Interaction fingerprints in chain A (PDB: 4ZQK\_B) calculated by MaSIF. Subfigures (a) and (b) are plotted using PyMol \cite{pymol}. Subfigures (c) and (d) are plotted using the PyMol surface plugin available in \cite{gainza2020}.}
\label{fig5}
\end{figure}

Figure~\ref{fig5} shows sites likely to interact, as predicted by MaSIF. The surface of the experimentally defined structure for the ligand-bound human protein, human programmed death-1 (PD-1) (PDB: 4ZQK), is presented in Fig.~\ref{fig5}(b) where it is possible to observe sections from each chain that interact in the protein complex. Visualizations of the MaSIF predictions are depicted in Figs.~\ref{fig5}(c) and (d). Figure~\ref{fig5}(c) shows regions on the surface of chain A that are most likely to interact, while Fig.~\ref{fig5}(d) displays the surface of chain B that is most likely to interact. The MaSIF model can identify critical interaction regions, and the visualization of the scores predicted by MaSIF is an effective method for analyzing interactions. Machine learning methods using interpretable labels such as MaSIF are gaining attention as they provide crucial feedback for researchers. Such ML models can predict site interactions and complement challenging techniques like docking methods. Recently, a method combining MaSIF surface features for docking has been shown to achieve state-of-the-art results for general proteins  \cite{sverrisson2023diffmasif}. Figure~\ref{fig5} highlights that the molecular surface, a high-level representation of protein structures, is a powerful visualization for interaction analysis.

\section{Discussion}\label{sec5}

The effective adoption of interpretable ML methods by non-ML experts depends on the use of powerful visualizations that convey information effectively to the user  (Section 3). To improve utility for protein applications, in addition to developing novel machine learning methods capable of giving interpretable explanations for their decisions, these explanations also need to be visualized effectively using new paradigms. Currently, complex visualizations for presenting protein structures and interpreting ML models are often designed manually for specific scenarios \cite{Chatzimparmpas2020}. Current advances include designing add-ons \cite{molecularnodes} to 3D creation software suites such as Blender \cite{blender} to provide a powerful protein structure representation. Improvements in creating complex and informative visualizations will be crucial to identifying high-order residue-residue interactions and recognizing the importance of high-dimensional features from different backgrounds, for example, combining pure structural features with the chemical properties of each residue.
 
In this survey, we focused on non-compressed representations (Section 2), however, various representation learning methods have been proposed to extract a compact representation of protein structures \cite{wu2022survey}. For these, effective ways to extract knowledge are needed. Intuitively, these methods are trained to learn a low-dimensional representation of the protein structure. These compact protein structure representations are computationally efficient, especially in handling large biomolecular and biomolecular complexes, and offer the possibility of interpreting protein structures in a lower dimensional latent space. In Zhang et al \cite{zhang2022} and Durairaj et al \cite{durairaj2021}, the learned latent space is mapped into two-dimensional space by UMAP \cite{mcinnes2018} and tSNE \cite{vandermaaten2008}, respectively, and clustered for further visualization. These methods show that learned latent spaces tend to be close in space for proteins from similar superfamilies while being far apart for proteins from different superfamilies \cite{zhang2022}. Visualizing characteristics of protein structures in low dimensional spaces can be an effective way to identify the functional capabilities of de novo proteins and improve sample efficiency using ML methods, such as reinforcement learning-based algorithms, for protein engineering and protein design.
 
For tasks predicting labels for each protein, the ability to interpret ML methods relies mostly on post-hoc methods (Table 1). Nonetheless, these methods have been criticized for giving unreliable explanations or being exploitable. In \cite{laugel2019} it is argued that for specific datasets, there is a high risk that the explanation provided is the result of artifacts learned by the model. Slack et al. \cite{slack2020} question the reliability of posthoc methods and find that methods such as LIME \cite{ribeiro2016} and SHAP \cite{lundberg2017} can be vulnerable to adversarial attacks. The trustworthiness of the use of saliency maps to accurately interpret biomedical images has also been questioned \cite{saporta2021, arun2021}. The concern has also been raised that attention weights may be unable to give reliable interpretations of graph structures \cite{ying2019, yu2021}. Finally, results obtained by reducing the latent space by Principal Component Analysis (PCA) and further clustering have been questioned since they can be difficult to reproduce for various genomic datasets \cite{elhaik2022}. Therefore, extensive studies on interpretable machine learning are needed, especially those that support the further development of inherently interpretable ML methods in which explainability is part of the architecture. The development of these methods will provide the ability to understand individual predictions, global feature interactions, and other crucial parameters, such as the landscape of the loss functions used to train these models.
 
Finally, when discussing interpretable ML models, it is important to develop a clear and rigorous definition of interpretability based on definitions from explainable/interpretable AI and interdisciplinary research (Section 1). In this paper, it is noted that we focus on interpretability to understand the decision-making process of ML methods following the definition given by Murdoch et al  \cite{murdoch2019}. These authors hypothesized that interpretable ML methods and effective visualizations are crucial to confirming existing knowledge and generating new assumptions in science. Here, we have further reinforced the idea that interpretations provided by the ML model should be understandable both by a designer with a background in ML and AI research and the researcher who will perform the experiments in the laboratory.
\section{Conclusion}\label{sec6}

Progress in the development of ML models using structure-based protein representations has led to novel architectures that are incipient in the biology and ML research communities. Significant advances from these architectures include novel modules for handling properties related to protein structures and adding new objective functions to enhance model training. Metrics such as the pLDDT in AF2 for measuring model confidence have been shown to provide important insights to the designer when combined with structure-based visualizations. Critically, these metrics have drastically enhanced the ability to filter potential candidates, leading to higher success rates in wet lab environments.

In this paper, we highlighted the importance of recent advances and discussed the current state of interpretable machine learning methods in computational biology. Despite advances in the field, there are still several limitations to be further explored. For example, various models do not include interpretable modules in their architecture and still rely on post hoc methods for interpretability. These limitations offer new opportunities for the development of inherent interpretable ML methods. We highlight that future models should consider not only accuracy metrics but also information that gives insights to users, i.e., which residues are important, which interactions are important, how confident the model is in its prediction, and its development process. For recent generative AI methods used for protein design~\cite{zhu2024generative, tang2024survey}, such as diffusion-based and graph-based methods, developing novel architectures with inherent interpretable modules and scores offer a promising research direction.

Additional limitations include current visualization tools to understand interpretable metrics, especially when these involve the interaction of multiple residues. Future work should focus on interdisciplinary collaborations so that interpretable modules can be effectively visualized and understood by structural biologists. These include the development of modules and tools to handle high-dimensional structural embeddings and metrics, which still rely on dimensionality reduction algorithms. In summary, developing interpretable models and novel visualization tools is critical for speeding up drug discovery by providing novel biological insights and enhancing the identification of important biological features and functional candidates in silico.






\bibliographystyle{ACM-Reference-Format}
\bibliography{interpretable-ml-manuscript}

\end{document}